# Gray level image enhancement using the Bernstein polynomials

Vasile P. Pătraşcu[1]

**Abstract** – This paper presents a method for enhancing the gray level images. This presented method takes part from the category of point operations and it is based on piecewise linear functions. The interpolation nodes of these functions are calculated using the Bernstein polynomials.
**Keywords:** image enhancement, piecewise linear function, Bernstein polynomials.

## I. INTRODUCTION

Image enhancement is an important task in the field of image processing. Lengthways time, there were built many methods in this purpose [1], [2], [6], [7]. The great number of existing methods is determined by the great variety of images, which need specific methods [4]. This paper describes a method of image enhancement, which takes part in the group of point transforms [3]. The piecewise linear functions are used for gray level transforms. These functions have a graphic represented by polygonal lines [5], while the interpolation nodes are determined using the Bernstein polynomials. The following part of the article is organized thus: section 2 comprises the mathematical theory presentation for the piecewise linear functions while the section 3 introduce the λ-means, the λ-histograms. The piecewise linear approximation for the gray level transform and the experimental results are presented in section 4. Finally the section 5 offers several conclusions.

## II. THE PIECEWISE LINEAR FUNCTIONS

Given the real numbers $v_1 < v_2 < ... < v_{n-1} < v_n$ let $f$ be a continuous function $f : R \to R$ that is linear on each interval $[v_i, v_{i+1}]$, $i = 1, 2, ..., n-1$. Such a function is called piecewise linear and the numbers $v_i$ are called breakpoints. Piecewise linear functions (PL) are often used to approximate (continuous) nonlinear functions. The PL are also known as linear splines with variable knots. The graphic of a PL function is represented with a polygonal line [5]. Usually the gray level set is a real bounded interval $E = [0, M]$ with $M \in (0, \infty)$. Because of that for image processing one considers the following particularly family of the PL:

$$f : [0, M] \to [0, M], \quad f(v) = \sum_{i=1}^{n} a_i |v - v_i| \qquad (1)$$

This function depends on the real parameters $(a_i)_{i=1,n}$ and $(v_i)_{i=1,n}$. Let be $f_1, f_2, ..., f_{n-1}, f_n$ a set of values from $[0, M]$. The elements $(f_i)_{i=1,n}$ are the function values in the breakpoints $(v_i)_{i=1,n}$. One determines the coefficients $a_1, a_2, ..., a_n$ from the conditions:

$$f(v_i) = f_i \quad \text{with } i = 1, 2, ..., n \qquad (2)$$

The system (2) has the subsequent solution [5]:

$$\begin{cases} a_1 = \frac{1}{2}\left(\frac{f_n + f_1}{v_n - v_1} + \frac{f_2 - f_1}{v_2 - v_1}\right) \\ a_i = \frac{1}{2}\left(\frac{f_{i+1} - f_i}{v_{i+1} - v_i} - \frac{f_i - f_{i-1}}{v_i - v_{i-1}}\right), i = 2, ..., n-1 \\ a_n = \frac{1}{2}\left(\frac{f_n + f_1}{v_n - v_1} - \frac{f_n - f_{n-1}}{v_n - v_{n-1}}\right) \end{cases} \qquad (3)$$

The relations (3) supply the coefficients values $(a_i)_{i=1,n}$ in determining the function $f$ when knowing the values $(f_i)_{i=1,n}$ in the points $(v_i)_{i=1,n}$.

## III. λ-MEANS AND λ-HISTOGRAM FOR GRAY LEVEL IMAGES

### A. *The Bernstein polynomials*

Polynomials are incredibly useful mathematical tools as they are simply defined and can be calculated quickly on computer systems. The most used form is:
$$P(t) = a_0 + a_1 t + ... + a_{k-1} t^{k-1} + a_k t^k$$
which represents a polynomial of degree $k$ as a linear combination of certain elementary polynomials $\{1, t, t^2, ..., t^k\}$. The set of polynomials of degree less then or equal to $k$ forms a vector space. The set of functions $\{1, t, t^2, ..., t^k\}$ forms a basis for this vector space. In this approach it use the Bernstein basis, which are different to the common bases for the space of polynomials. The Bernstein polynomials of degree $k$ are defined by: $B_{ik}: [0, M] \rightarrow [0,1]$,

$$B_{ik}(t) = C_k^i \left(\frac{t}{M}\right)^i \left(1 - \frac{t}{M}\right)^{k-i} \quad \text{for } i = 0, 1, ..., k \quad (4)$$

where $C_k^i = \frac{k!}{i!(k-i)!}$. The $k+1$ Bernstein polynomials of degree $k$ form a partition of unity:

$$\sum_{i=0}^{k} B_{ik}(t) = 1 \quad (5)$$

[1] TAROM Company,
  Department of Informatics Technology,
  e-mail: vpatrascu@tarom.ro

B. *The $\lambda$-means for gray level images*

A gray level image is defined on a compact spatial domain $D \subset R^2$ by a gray level function $l: D \rightarrow E$ where $E = [0, M]$ is the gray level set. Usually $M = 255$. Assume that the function $l$ is a continuous one.

The Bernstein means $b_0(l), b_1(l), ..., b_k(l)$ are defined as following: for $i = 0, 1, ..., k$

$$b_i(l) = \frac{\int_D B_{ik}(l(x,y)) \cdot l(x,y) dx dy}{\int_D B_{ik}(l(x,y)) dx dy} \quad (6)$$

It can use a $\lambda$ tuning parameter to obtain more flexibility. Let be the functions: $F_{ik}: [0, M] \rightarrow [0,1]$,

$$F_{ik}(t) = \frac{\left(B_{ik}(t)\right)^\lambda}{\sum_{i=0}^{k} \left(B_{ik}(t)\right)^\lambda} \quad \text{for } i = 0, 1, ..., k \quad (7)$$

where $\lambda \in (1, \infty)$.

From (7) it results a similarly property like (5):

$$\sum_{i=0}^{k} F_{ik}(t) = 1 \quad (8)$$

With the functions (7) one can define the $\lambda$-means: for $i = 0, 1, ..., k$

$$b_{\lambda, i}(l) = \frac{\int_D F_{ik}(l(x,y)) \cdot l(x,y) dx dy}{\int_D F_{ik}(l(x,y)) dx dy} \quad (9)$$

C. *The $\lambda$-histograms for gray level images*

Let be a gray level image $l: D \rightarrow E$ where $l$ is continuous function. The Bernstein histogram is a discrete one. Each bin $h_i(l)$ is given by:

$$h_i(l) = \frac{\int_D B_{ik}(l(x,y))}{area(D)} \quad \text{for } i = 0, 1, ..., k \quad (10)$$

Using the functions $F_{ik}$ defined by (7) one can define the $\lambda$-histograms:

$$h_{\lambda, i}(l) = \frac{\int_D F_{ik}(l(x,y))}{area(D)} \quad \text{for } i = 0, 1, ..., k \quad (11)$$

which belongs also to the discrete histograms family. It is very simple to prove that the following relations result from (10) and (11):

$$\sum_{i=0}^{k} h_i(l) = 1,$$

$$\sum_{i=0}^{k} h_{\lambda, i}(l) = 1$$

Using the bins of $\lambda$-histogram (11) we can define the accumulated $\lambda$-histogram as follows:

$$H_{\lambda, 0}(l) = \frac{1}{2} \cdot h_{\lambda, 0}(l) \quad (12a)$$

$$\begin{cases} H_{\lambda, i}(l) = H_{\lambda, i-1}(l) + \frac{1}{2}\left(h_{\lambda, i}(l) + h_{\lambda, i-1}(l)\right) \\ \text{for } i = 1, ..., k \end{cases} \quad (12b)$$

IV. THE ENHANCEMENT PROCEDURE

A. *The gray level transform for image enhancement*

Let be $l: D \rightarrow E$ the image that must be enhanced and $u: D \rightarrow E$ an image with an uniform distribution of gray levels. The function that defines the gray level transform is a PL one with $n$ nodes. For the interpolation nodes are used the $\lambda$-means and the bins of the accumulated $\lambda$-histogram computed with relations (9) and (12) for $k = n - 3$. The breakpoints $v_1$ and $v_n$ are steady in the values: $v_1 = \min_{(x,y) \in D} l(x,y)$, $v_n = \max_{(x,y) \in D} l(x,y)$. The others $n - 2$ breakpoints $v_2, v_3, ..., v_{n-1}$ will be the $\lambda$-means $b_{\lambda, 0}(l), b_{\lambda, 1}(l), ..., b_{\lambda, k}(l)$. So, $v_{i+2} = b_{\lambda, i}(l)$, for $i = 0, 1, ..., k$. To define the PL function for image enhancement we need the parameters $(\alpha_i)_{i=0,k}$ and $\beta$ that are defined by:

$$\alpha_i = \frac{H_{\lambda, i}(l)}{H_{\lambda, i}(u)} \quad \text{for } i = 0, 1, ..., k \quad (13)$$

and

$$\beta = \frac{\alpha_0 \cdot b_{\lambda, 0}(u) + \sum_{i=1}^{k} \alpha_i \left(b_{\lambda, i}(u) - b_{\lambda, i-1}(u)\right)}{M} \quad (14)$$

The values $(f_i)_{i=1,n}$ of the interpolation function are computed with the following relations:

$$f_1 = 0, \quad f_2 = \frac{\alpha_0}{\beta} \cdot b_{\lambda,0}(u) \tag{15a}$$

$$\begin{cases} f_i = f_{i-1} + \frac{\alpha_{i-2}}{\beta}\left(b_{\lambda,i-2}(u) - b_{\lambda,i-3}(u)\right) \\ \text{for } i = 3,..,n-1 \end{cases} \tag{15b}$$

$$f_n = f_{n-1} + \frac{1}{\beta}\left(M - b_{\lambda,n-1}(u)\right) \tag{15c}$$

Having the values function $(f_i)_{i=1,n}$ in the points $(v_i)_{i=1,n}$, we can compute the coefficients values $(a_i)_{i=1,n}$ using (3). Then applying the PL function $f$ to the original image $l$, we obtain the enhanced image $l_{enh} = f(l)$.

B. *The enhancement procedure*

Let be $l: D \to E$ the image that must be enhanced and $u: D \to E$ an image with an uniform distribution of gray levels. The next procedure will be used for enhanced image calculus:

1. Initialization: choose $k, \lambda, \varepsilon$ (the constant for stopping the procedure) and $m = 0$. We set $l^{(0)} = l$ and compute the $\lambda$-means $\left(b_{\lambda,i}(l^{(0)})\right)_{i=0,k}, \left(b_{\lambda,i}(u)\right)_{i=0,k}$ and the accumulated $\lambda$-histograms $\left(H_{\lambda,i}(l^{(0)})\right)_{i=0,k}, \left(H_{\lambda,i}(u)\right)_{i=0,k}$ for the images $l^{(0)}$ and $u$.

2. We compute the parameters $\left(\alpha_i^{(m)}\right)_{i=0,k}, \beta^{(m)}$ and the values $\left(f_i^{(m)}\right)_{i=1,n}$ using the relations (13, 14, 15). One calculate $v_1^{(m)} = \min_{(x,y) \in D} l^{(m)}(x,y)$, $v_{k+3}^{(m)} = \max_{(x,y) \in D} l^{(m)}(x,y)$ and $v_{i+2}^{(m)} = b_{\lambda,i}(l^{(m)})$ for $i = 0,1,...,k$. One compute the piecewise linear function $f^{(m)}:[0,M] \to [0,M]$,

$$f^{(m)}(v) = \sum_{i=1}^{k+3} a_i^{(m)} \left| v - v_i^{(m)} \right| \text{ using relations (3),}$$

$$a_1^{(m)} = \frac{1}{2}\left( \frac{f_{k+3}^{(m)} + f_1^{(m)}}{v_{k+3}^{(m)} - v_1^{(m)}} + \frac{f_2^{(m)} - f_1^{(m)}}{v_2^{(m)} - v_1^{(m)}} \right)$$

$$\begin{cases} a_i^{(m)} = \frac{1}{2}\left( \frac{f_{i+1}^{(m)} - f_i^{(m)}}{v_{i+1}^{(m)} - v_i^{(m)}} - \frac{f_i^{(m)} - f_{i-1}^{(m)}}{v_i^{(m)} - v_{i-1}^{(m)}} \right) \\ \text{for } i = 2,...,k+2 \end{cases}$$

$$a_{k+3}^{(m)} = \frac{1}{2}\left( \frac{f_{k+3}^{(m)} + f_1^{(m)}}{v_{k+3}^{(m)} - v_1^{(m)}} - \frac{f_{k+3}^{(m)} - f_{k+2}^{(m)}}{v_{k+3}^{(m)} - v_{k+2}^{(m)}} \right)$$

We will compute $l^{(m+1)} = f^{(m)}(l^{(m)})$, the $\lambda$-means $\left(b_{\lambda,i}(l^{(m+1)})\right)_{i=0,k}$, the accumulated $\lambda$-histogram $\left(H_{\lambda,i}(l^{(m+1)})\right)_{i=0,k}$ using (9), (12) for this new image. To obtain the finally gray level transform $\psi_{enh}$ we must compute the string of functions $\psi^{(m+1)} = \psi^{(m)} \circ f^{(m)}$ where $\psi^{(0)} = 1_E$ is the identity function on gray level set $E$.

3. If $\left| b_{\lambda,i}(l^{(m+1)}) - b_{\lambda,i}(u) \right| < \varepsilon$ pass to the step 4, otherwise $m = m+1$ and go to step 2.

4. Save the results $l_{enh} = l^{(m+1)}$, $\psi_{enh} = \psi^{(m+1)}$, $\left(b_{\lambda,i}(l^{(m+1)})\right)_{i=0,k}, \left(h_{\lambda,i}(l^{(m+1)})\right)_{i=0,k}$ and stop.

To exemplify, three images were picked out: one dark ("tire") in Fig.1a, one bright ("cells") in Fig.2a. and one with low contrast ("lax") in Fig.3a. Their $\lambda$-histograms are in Fig.1b, Fig.2b and Fig.3b. The graphics of their gray level transforms $\psi_{enh}$ are shown in Fig.1c, Fig.2c and Fig.3c. The enhanced images can be seen in Fig.1d, Fig.2d and Fig.3d with the $\lambda$-histograms in Fig.1e, Fig.2e and Fig.3e. The $\lambda$-means for the original images and for the enhanced images are represented in Fig.1f, Fig.2f and Fig.3f.

V. CONCLUSIONS

The paper presented a method for enhancing the gray level images. The method is based on point transforms defined by piecewise linear functions. These functions are determined by simple formulae, which need short calculus time. In establishing the interpolation points it was chosen an algorithm that is similarly to the classical histogram equalization. Future perspectives for the shown method could be the extension for color images.

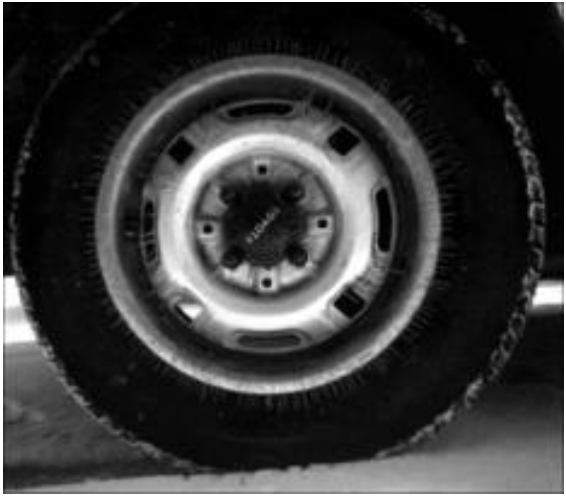

Fig. 1. a) The original image "tire"

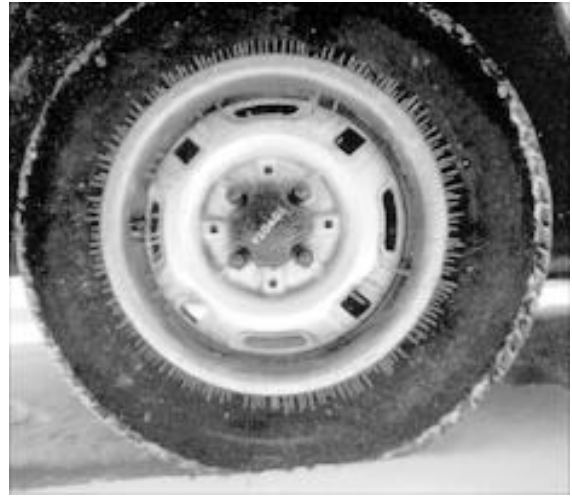

Fig. 1. d) The enhanced image

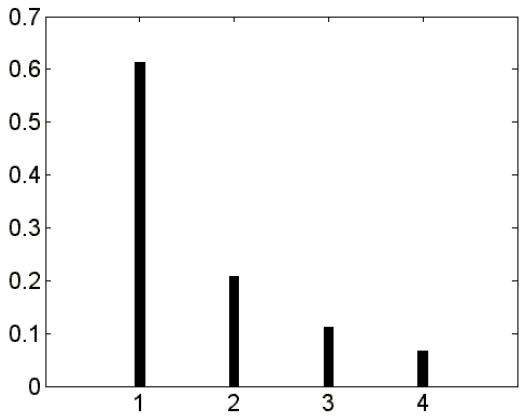

Fig. 1. b) The λ - histogram of original image

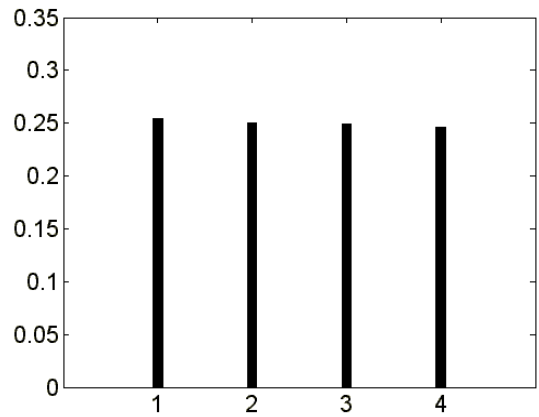

Fig. 1. e) The λ - histogram of enhanced image

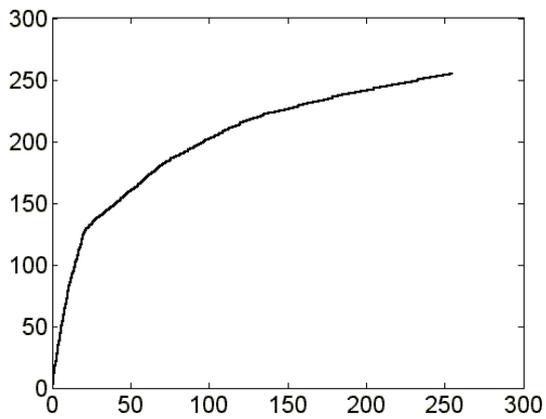

Fig. 1. c) The gray level transform

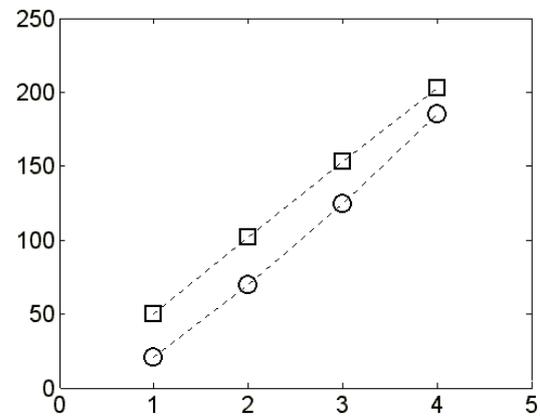

Fig. 1. f) The λ - means for original image (circle) and for enhanced image (square)

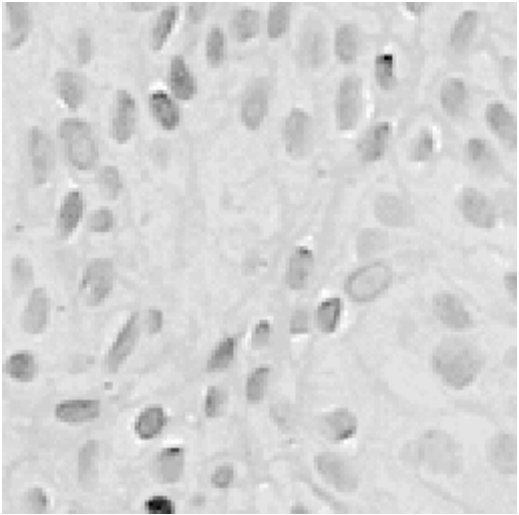

Fig. 2. a) The original image "cells"

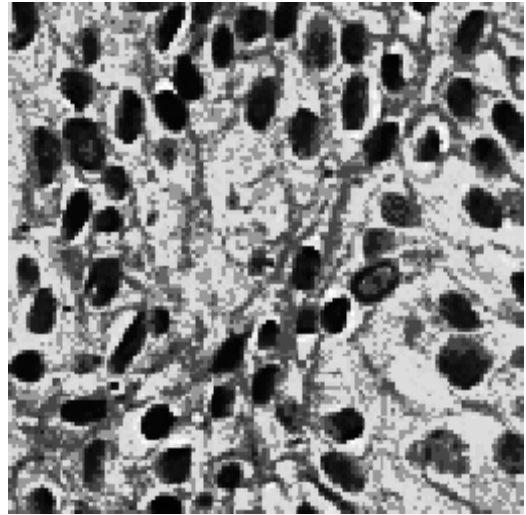

Fig. 2. d) The enhanced image

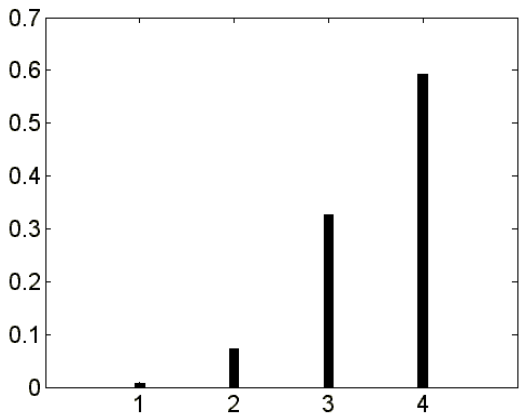

Fig. 2. b) The λ - histogram of original image

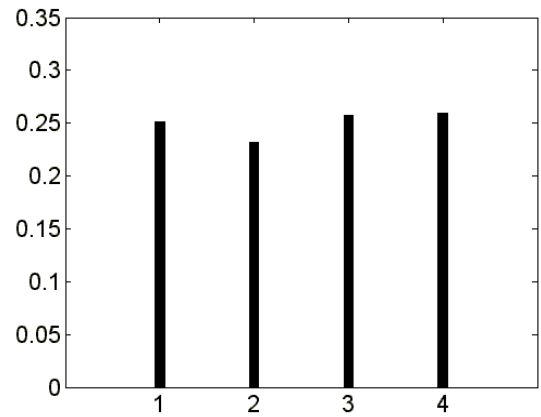

Fig. 2. e) The λ - histogram of enhanced image

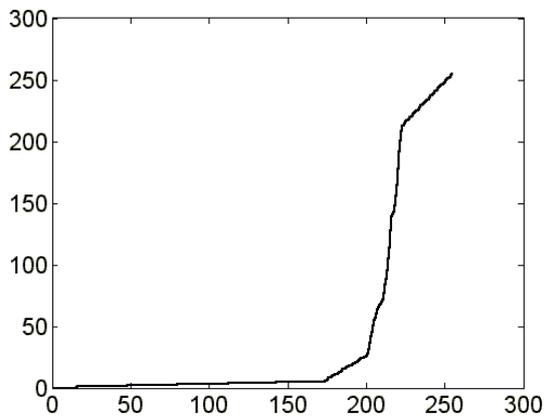

Fig. 2. c) The gray level transform

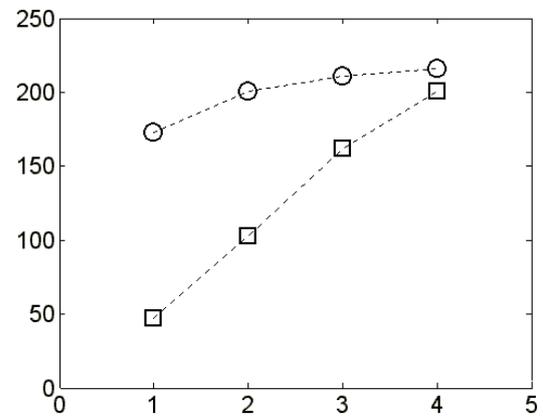

Fig. 2. f) The λ - means for original image (circle) and for enhanced image (square)

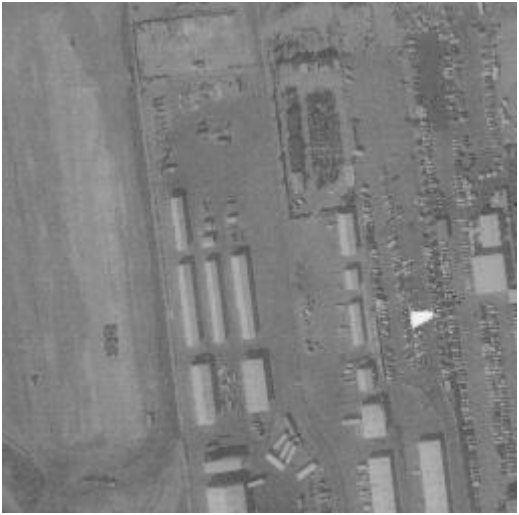

Fig. 3. a) The original image "lax"

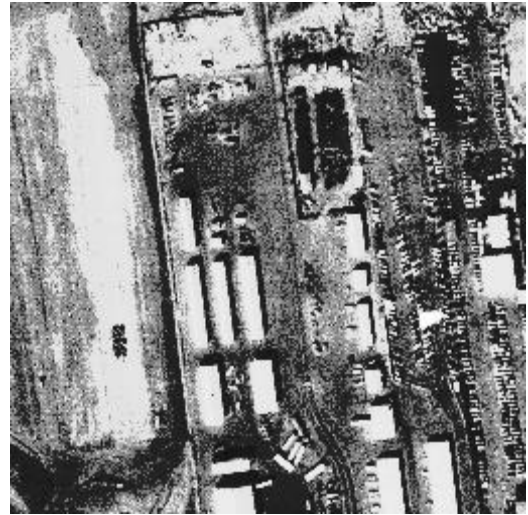

Fig. 3. d) The enhanced image

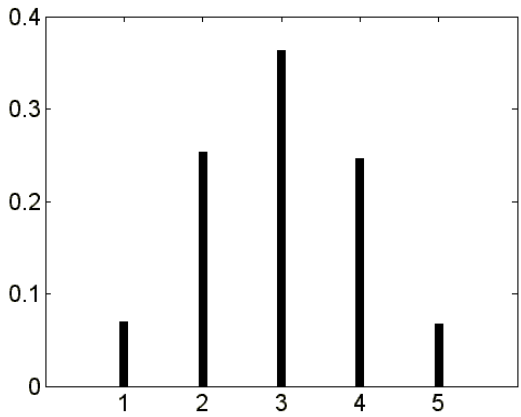

Fig. 3. b) The $\lambda$ - histogram of original image

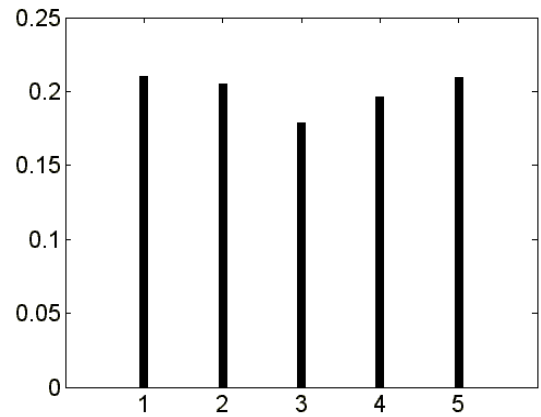

Fig. 3. e) The $\lambda$ - histogram of enhanced image

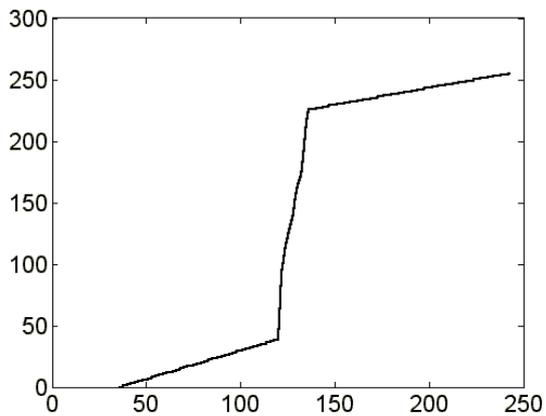

Fig. 3. c) The gray level transform

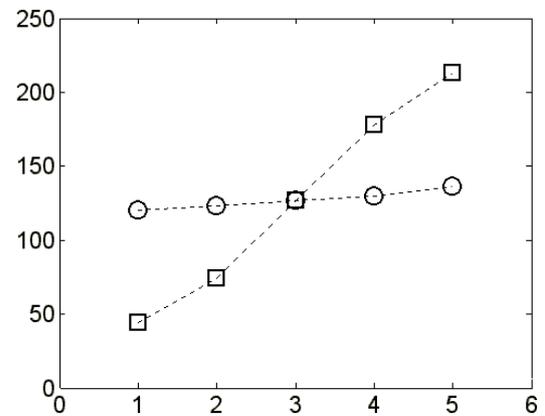

Fig. 3. f) The $\lambda$ - means for original image (circle) and for enhanced image (square)